\documentclass[11pt,a4paper]{article}
\usepackage[hyperref]{emnlp2018}
\usepackage{times}
\usepackage{latexsym}

\usepackage{url}
\usepackage{float}
\usepackage{graphicx}
\usepackage{amsmath}
\usepackage{amssymb}
\usepackage{bm}

\usepackage{multirow}

\aclfinalcopy 

\newcommand*{\affaddr}[1]{#1} 

\newcommand*{\email}[1]{\texttt{#1}}

\title{Learning Sentiment Memories for Sentiment Modification \\
without Parallel Data}

\author{
Yi Zhang, \ 
Jingjing Xu, \ 
Pengcheng Yang, \
Xu Sun
\\ 
\affaddr{MOE Key Lab of Computational Linguistics, School of EECS, Peking University}  \\
\email{\{zhangyi16,jingjingxu,yang\_pc,xusun\}@pku.edu.cn }\\
}

\date{}

\begin{document}
\maketitle
\begin{abstract}
The task of sentiment modification requires reversing the sentiment of the input and preserving the sentiment-independent content. However, aligned sentences with the same content but different sentiments are usually unavailable. Due to the lack of such parallel data, it is hard to extract sentiment independent content and reverse the sentiment in an unsupervised way. Previous work usually can not reconcile sentiment transformation and content preservation. In this paper,
motivated by the fact the non-emotional context (e.g., ``staff") provides strong cues for the occurrence of emotional words (e.g., ``friendly"), we propose a novel method that automatically extracts appropriate sentiment information from learned sentiment memories according to specific context. Experiments show that our method substantially improves the content preservation degree and achieves the state-of-the-art performance. \footnote{The code is available at \url{https://github.com/lancopku/SMAE}} 
\end{abstract}

\setlength{\abovedisplayskip}{2pt}
\setlength{\abovedisplayshortskip}{2pt}
\setlength{\belowdisplayskip}{2pt}
\setlength{\belowdisplayshortskip}{2pt}

\section{Introduction}
Sentiment modification of natural language texts is a special task that connects sentiment analysis and natural language generation. It facilitates many NLP applications, such as news rewriting and automatic conversion of review attitude, which reduce the human effort. Sentiment modification presents two requirements: one is that the sentiment or the attitude of the text needs to be transformed to the opposite; the other is that the transformed text should maintain semantic relevance to the input text as much as possible.


Recently, there have been some researches which focus on the work of editing a sentence to alter specific attributes, like style and sentiment \cite{Shen17crossalign, Hu17controlled}. Typically, the parallel data with the same content but different sentiment is usually not available. This line of work attempts to extract the attribute-independent content from a dense sentence representation by adversarial learning. However, it is hard to extract the attribute-independent content in such implicit ways, which makes these methods tend to generate input-irrelevant texts.


Most existing methods can not reconcile the performance of sentiment transformation and content preservation. Direct replacement of emotional words can keep the context but may lead to low-quality sentences. For example, given an input ``\emph{The food is cold like rock}'', this method probably outputs ``\emph{The food is warm like rock}''. State-of-the-art models using neural networks struggle to generate high-quality sentences. However, these models usually lead to poor content preservation.  For instance, when the source text is ``\emph{This is a wonderful movie}", we expect an output like ``\emph{This movie is disappointing}". However, the generated sentence may be ``\emph{The waiters are very rude}", which has little relevance with the source text. In general, it is difficult to preserve semantic content and reverse the sentiment at the same time without parallel data.



\begin{figure}[t] 
\centering
\includegraphics[width = 0.9\linewidth]{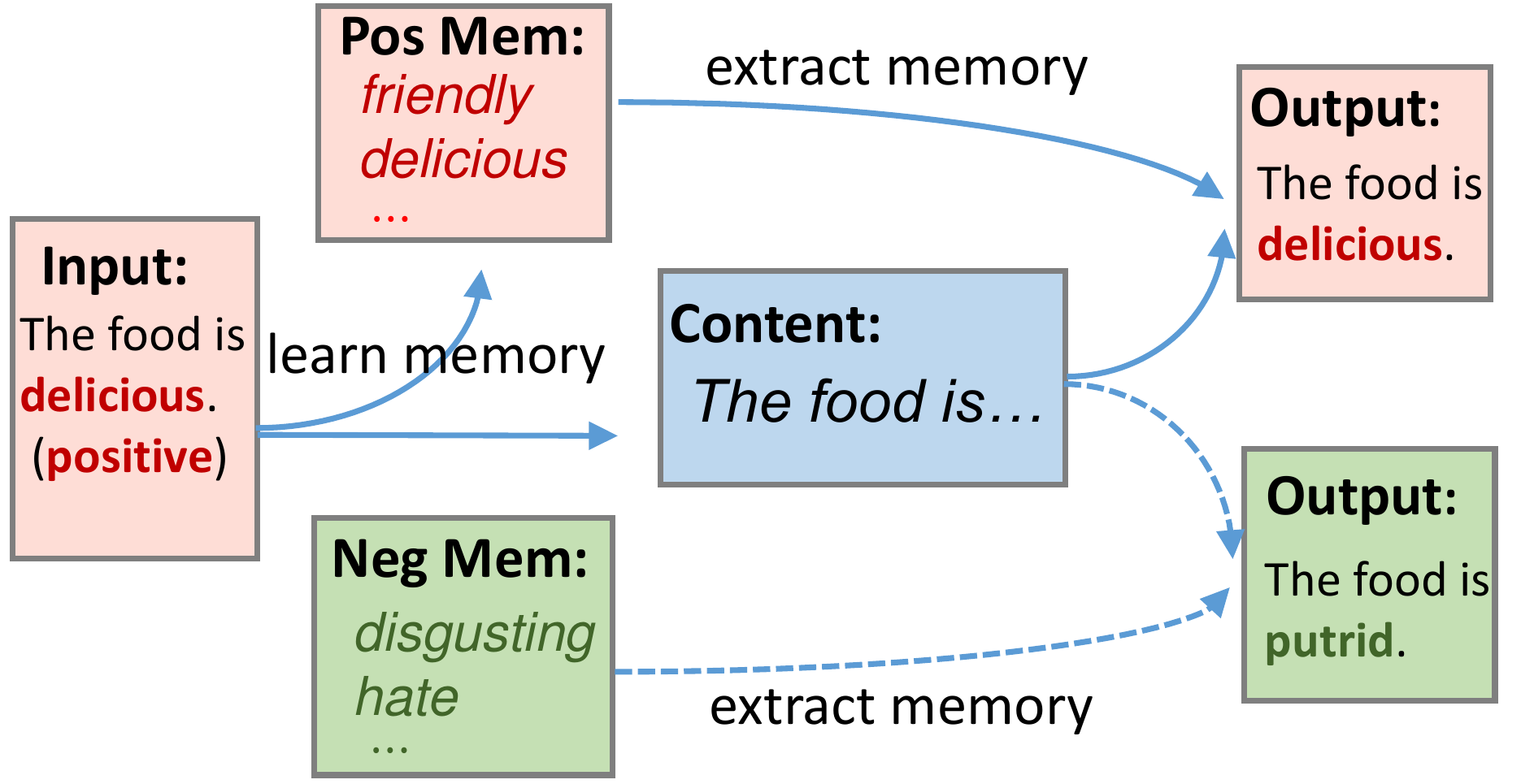}
\caption{Illustration of the proposed model with a positive input. Solid and dashed lines indicate the training process and the testing process, respectively. The process with a negative input is in a similar way.}
\label{structure} 
\end{figure}

To address this problem, we propose a novel model which performs well in both sentiment transformation and content preservation. Our model first learns two kinds of sentiment memories by explicitly separating emotional words. Then, according to the specific context, the model extracts appropriate sentiment information from the memory of target sentiment. The decoder takes the extracted memory and the context representation together to perform decoding. The overview of our model is shown in Figure~\ref{structure}.
The main architecture of our model is a 
\textbf{S}entiment-\textbf{M}emory based \textbf{A}uto-\textbf{E}ncoder (\textbf{SMAE}). The proposed model achieves the state-of-the-art performance, especially improves content preservation degree.

\setlength{\abovedisplayskip}{2pt}
\setlength{\abovedisplayshortskip}{2pt}
\setlength{\belowdisplayskip}{2pt}
\setlength{\belowdisplayshortskip}{2pt}


Our contributions are concluded as follows:
\begin{itemize}
\item We propose a method that uses sentiment memories to accomplish sentiment modification without any help of the parallel data.
\item The proposed method improves the content preservation degree by a large margin when compared with current systems.
\end{itemize}

\section{Related Work}

Recently, there has been some studies for sentiment modification. \newcite{Shen17crossalign} learn an encoder that maps a sentence with its original style to a style-independent content representation. This is then passed to a style-dependent decoder for rendering. \newcite{zhenxinStyTranfer} implement a multi-decoder auto-encoder \cite{bengio2009learning,dai2015semi} where the encoder is used to capture the content and the sentiment-specific decoders are used to generate target sentence. \newcite{Hu17controlled} augment the unstructured variables $z$ in vanilla VAE with a set of structured variables $c$ each of which targets a salient and independent semantic feature of sentences, to control sentence sentiment. However, all of these work attempt to implicitly separate the non-emotional content from the emotional information in a dense sentence representation.  \newcite{xu2018unpaired} explicitly filter out emotional words. They use two sentiment-specific decoders to attach sentiments to non-emotional context. The decoders bear all the burdens to generate sentiments. In our model, we use sentiment memories to assist generating sentiments with only one decoder, results in fewer parameters.

The proposed sentiment-memory based auto-encoder \cite{bengio2009learning,P18-2115} learns the idea of memory network \cite{memory_WestonCB14, mmSukhbaatarSWF15} but simplifies the process. Our work is also related to the generation tasks  \cite{DBLP:conf/emnlp/WangLSC17, DBLP:conf/aaai/LiuWSCS18, DBLP:journals/corr/abs-1805-01089,P18-2027}. These tasks usually generate texts that preserve main information of input texts.

\section{Proposed Model}
We first use a variant of self-attention\cite{DBLP:journals/corr/LinFSYXZB17,KimDHR17_attention} mechanism to distinguish the emotional and non-emotional words. Then the positive words and negative words are used to update the corresponding memory modules. Finally, the decoder uses the target sentiment information extracted from the memory and the content representation to perform decoding.

\subsection{Emotional Words Detection Model}
We first find the emotional words that have the most discriminative power for sentiment polarity. This work is done by training a sentiment classifier with a simple self-attention mechanism. Here the sequence of inputs $\{h_1, . . . , h_T\}$ are the hidden states of a LSTM, running over the words in the source sentence $\{x_1, . . . , x_T\}$. The context vector can then be computed using a simple sum:
\begin{align}
\boldsymbol{c} &= \sum_{t=1}^{T}a_{t} \cdot \boldsymbol{h}_t 
\end{align}
where $a_t$ denotes the attention weight of the $t$-th word. The sentence vector $\boldsymbol{c}$ is then fed into a fully connected layer to predict the sentiment polarity of the source text. Since the words with obvious emotional tendencies will be given greater weights compared to those non-emotional words during training, $a_t$ can be used to distinguish between emotional and non-emotional words.

The weights of standard attention mechanism sum to 1. When there are several emotional words, the sum 1 is distributed by these words. However, we expect that each emotional word has a weight close to 1 to identify its sentiment attribute. Hence, following \cite{KimDHR17_attention}, we modify the calculation of attention weights as follows to get more distinguishable weights:
\begin{equation}
a_t = {\rm sigmoid} (\boldsymbol{v}^T  \boldsymbol{h}_t)
\end{equation}
where $\boldsymbol{v}$ is the parameter vector. 
The $\rm sigmoid$ function follows our intention that giving each input word a distinguishable weight which is close to 1 or 0. 
However, these weights falls between 0 and 1. They still can not thoroughly distinguish the emotional words from non-emotional words without redundant information. Following \newcite{xu2018unpaired}, we map attention
weights to discrete values, 0 or 1, and we adopt their discrete method. The weights greater than the averaged attention value are assigned to 1 and the weights less than the averaged attention value are assigned to 0. The weight $a_t$ after discretization is denoted as $\hat{a_t}$. Then, $\hat{a_t}$ can be regarded as the emotional word identifier. $1-\hat{a_t}$ becomes non-emotional word identifier.

\subsection{Sentiment-Memory Based Auto-Encoder}
After the separation of emotional and non-emotional words, the proposed \textbf{SMAE} is used to process these two kinds of information. We employ the seq2seq based auto-encoder. Both the encoder and the decoder are LSTM networks \cite{DBLP:journals/neco/HochreiterS97}. 

If $\boldsymbol{x}_i$ is a context word, then $\hat{a_i}$ is 0, causing $(1-\hat{a_i})\boldsymbol{x}_i$ to be $\boldsymbol{x}_i$. 
Therefore, the sequence $\{(1-\hat{a_1})\boldsymbol{x}_1, \cdots, (1-\hat{a_T})\boldsymbol{x}_T\}$ can be regarded as non-emotional word embedding sequence. It is fed into the LSTM encoder sequentially. we select $\boldsymbol{h}_T$ in the last state tuple ($\boldsymbol{h}_T, \boldsymbol{c}_T$) of the encoder as the content representation of the input.

 
Meanwhile, the embeddings of the emotional words of the source text are used to update the sentiment-memory. Since we have two kinds of sentiments, positive and negative, we use $\boldsymbol{M}^{pos} \in \mathbb{R}^{e \times \gamma}$ and $\boldsymbol{M}^{neg} \in \mathbb{R}^{e \times \gamma}$ to denote the positive memory and the negative memory, respectively. $e$ is the embedding size and $\gamma$ is a hyper-parameter which controls the size of the memory. 

We illustrate the following part using positive input as an example. We first sum the embedding of the emotional words to get a vector representation of the emotional information, which is denoted as $\boldsymbol{s}^{pos}\in \mathbb{R}^{e}$. We then use a simple attention mechanism to find the columns in $\boldsymbol{M}^{pos}$ that are most closely related to the emotional information. The outer product of the transposition of emotional information $\boldsymbol{s}^{pos}$ and the attention weights $\boldsymbol{w}$ broadcasts the sentiment vector $\boldsymbol{s}^{pos}$ to a matrix. Then, the matrix is added to the existing memory $\boldsymbol{M}^{pos}$. Due to the attention weight $\boldsymbol{w}$, the columns that are most closely related to the emotional information are updated more with the sentiment information $\boldsymbol{s}^{pos}$. 
%
Formally, we have:
\begin{align}
& \boldsymbol{s}^{pos} = \sum_{i = 1}^{T} \hat{a_i} \cdot \boldsymbol{x}_i \\
& \boldsymbol{w} = softmax\left((\boldsymbol{s}^{pos})^T  \boldsymbol{M}^{pos}\right) \\
& \boldsymbol{M}^{pos} = \boldsymbol{M}^{pos} +  {\boldsymbol{s}^{pos}} \otimes  \boldsymbol{w}
\end{align}
where $\otimes$ denotes the outer product.

Previous work employ two sentiment-specific decoders to generate text based on the supposed non-emotional representation. The decoders bear all the burdens to generate sentiments. In our model, we extract some sentiment information from the sentiment-memories to assist decoding. Intuitively, the context word ``staff'' is more likely to be associated with the emotional word ``friendly'', and ``food'' is more likely to be associated with ``delicious''. So we use the context vector $\boldsymbol{s}^{con}$ to extract the corresponding sentiment memory that is more likely to be used in the future decoding. The context vector $\boldsymbol{s}^{con}$ is represented as the sum of the embedding of non-emotional words. Then $\boldsymbol{s}^{con}$ is used to compute the attention weights $\boldsymbol{u}$ over the columns of sentiment memory matrix. We sum these weighted columns as the extracted memory $\tilde{\boldsymbol{m}}$ and add $\tilde{\boldsymbol{m}}$ to the last cell state $\boldsymbol{c}_T$ of the encoder:
\begin{align}
& \boldsymbol{s}^{con} = \sum_{i=1}^{T} (1 - \hat{a_i}) \cdot \boldsymbol{x}_i \\
& \boldsymbol{u} = softmax\left((\boldsymbol{s}^{con})^T  \boldsymbol{M}^{pos}\right)\\ 
& \tilde{\boldsymbol{m}} = \sum_{j=1}^{\gamma} u_j \cdot \boldsymbol{M}_j^{pos}\\
& \tilde{\boldsymbol{c}}_T = \boldsymbol{c}_T + W \tilde{\boldsymbol{m}} 
\end{align}
where $u_j$ denotes the $j$-th value in vector $\boldsymbol{u}$, $\boldsymbol{M}^{pos}_j$ denotes the $j$-th column of $\boldsymbol{M}^{pos}$ and $W$ is the parameter matrix. The new tuple $(\boldsymbol{h}_T, \tilde{\boldsymbol{c}}_T)$ then acts as the initial state of the decoder. 

The negative input is processed in the same way. At the training stage, the decoder is encouraged to restore the source text. Therefore, the cross entropy loss function is optimized.
\section{Experiments}    

\subsection{Data Preprocessing}
We use the Yelp Review Dataset (Yelp) provided by Yelp Dataset Challenge\footnote{\url{https://www.yelp.com/dataset/challenge}} to conduct experiments. Each item is a sentence from the review on Yelp and is labeled as having either negative or positive sentiment. We train a CNN sentence classifier \cite{CNN_YoonKim} to filter examples with ambiguous sentiment polarities (category probability $<$ 0.8). The processed dataset contains 510K, 20K, and 20K pairs for training, validation, and testing, respectively. The classifier achieves an accuracy of 94\% on the processed dataset and is also used to test transformation accuracy. 
 


\subsection{Experiment Settings}
We tune our hyper-parameters on the development set. The word embeddings are initialized randomly with a size of 128. The hidden size of the sentiment-memory based auto-encoder is 300. We use Adam optimizer \cite{Adam14} with a initial learning rate set to 0.001 to train our model and the batch size is set to 64. The hyper-parameter $\gamma$ which controls the size of memory matrix is 60.

\subsection{Baselines}
We compare our proposed method with two state-of-the-art systems that have been used for sentiment modification. We run the released code on our dataset.

\noindent\textbf{Cross-aligned Auto-Encoder (CAE)}: This system, proposed by~\citet{Shen17crossalign}, uses a shared latent content space across different sentiments and leverages refined alignment of latent representations to perform sentiment modification.

\noindent\textbf{Multi-decoder Auto-Encoder (MAE)}: This system is proposed by \citet{zhenxinStyTranfer}. They use a multi-decoder seq2seq model \cite{bengio2009learning, dai2015semi} where the encoder captures content information by adversarial learning \cite{goodfellow2014generative} and the sentiment-specific decoders are used to generate target sentences.
\begin{table}[t]
\setlength{\tabcolsep}{4pt}
\centering
    \begin{tabular}{l|l|l}
    \hline
    Model & ACC & BLEU    \\\hline
    CEA &71.96  &2.77  \\
    MAE &74.59   &5.45 \\ \hline
    \textbf{SMAE} & \textbf{76.64} (+2.05)& \textbf{24.00} (+18.55) \\\hline 
    \end{tabular}
    \caption{Performance of the proposed method and state-of-the-art systems.}
    \label{tab:result}
\end{table}

\subsection{Results and Discussions}
We use ACC to denote the transformation accuracy. Following \newcite{GanGHGD17}, we also compute BLEU \cite{papineni2002bleu} between the output and the source text to evaluate the content preservation degree. A high BLEU score primarily indicates that the system can correctly preserve content by retaining the same words from the source sentence.

\begin{table}[t]
\setlength{\tabcolsep}{4pt}
\centering
    \begin{tabular}{l|c|c|c}
    \hline
    Model & Sentiment & Content  & Fluency\\ \hline
    CAE &6.55  &4.46  &5.98 \\
    MAE &6.64  &4.43 &5.36 \\ \hline
  \textbf{SMAE} &6.57  &5.98  &6.69 \\
   \hline
    \end{tabular}
    \caption{Results of human evaluation.}
    \label{human_eval}
\end{table}

\begin{table}[t]
\centering
    \begin{tabular}{p{0.95\linewidth}}
    \hline
 
     \textbf{Input}: \textsl{Very helpful and informative staff!}\\
       \textbf{CAE}: \textsl{Worst service ever.}\\
       \textbf{MAE}:  \textsl{Very nice here and poor!}    \\    
       \textbf{Proposed}: \textsl{Very rude and careless staff !}\\ 
     \hline
    \textbf{Input}: \textsl{I will never go here again.}\\
       \textbf{CAE}: \textsl{I love this place here!}\\
       \textbf{MAE}:  \textsl{I had say this place here.}    \\    
       \textbf{Proposed}: \textsl{I will never go anywhere else.}\\  
    \hline
    
      \textbf{Input}: \textsl{The worst and would never recommend anyone to use them.}\\
       \textbf{CAE}: \textsl{The best place I 've been to go here!}\\
       \textbf{MAE}:  \textsl{The first experience is so happy and nice.}    \\    
       \textbf{Proposed}: \textsl{The best and would definitely recommend anyone to use them.}\\  
       \hline
    \end{tabular}
    \caption{Examples generated by the proposed method and baselines. In comparison, our model changes the sentiment of inputs with higher semantic relevance.}
    \label{samplecases}
\end{table}

The experimental results of our proposed model and the baselines are shown in Table~\ref{tab:result}. Both baseline models have low BLEU score but high accuracy, which indicates that they may be trapped in a situation that they simply output a sentence with the target sentiment regardless of the content. The main reason is that these methods using adversarial learning attempt to implicitly separate the emotional information from the context information in a sentence vector. However, without parallel data, it is difficult to achieve such a goal. Our proposed SMAE model takes advantage of self-attention mechanism and explicitly removes the emotional words, leading to a significant improvement of content preservation and the state-of-the-art performance in terms of both metrics. 


We also involve human evaluation to measure the quality of generated text. Each item contains an input and three outputs generated by different systems. Then 200 items are distributed to 2 annotators with linguistic background. The annotators have no idea about which system the output is from. They are asked to score the output on three criteria on a scale from 1 to 10: the transformed sentiment degree, the content preservation degree, and the fluency. Table~\ref{human_eval} shows the evaluation results. Our model has obvious advantage over the baseline systems in content preservation, and also performs well in other aspects. 

Several randomly selected examples generated by different models are shown in Table~\ref{samplecases}. These examples clearly show our proposed model can generate sentences that are more semantically relevant to the input text compared to the baselines.

\subsection{Effectiveness of Sentiment-Memories}

To verify the effectiveness of the memory module of our model, we conduct ablation study by excluding the sentiment-memory module. The result is shown in Table~\ref{tab_ablation}. According to the result, the complete model achieves an improvement of 62.56\% on transformation accuracy over the model that excludes the sentiment memories, which means the sentiment memories are key components to ensure successful sentiment modification. In addition, several examples are shown in Table~\ref{mem_examples} to visually demonstrate the effectiveness of the memory module. we can find that the proposed model is capable of generating appropriate emotional words (red words in Table~\ref{mem_examples}) to adapt different contexts. 

\begin{table}[tb] 
    \setlength{\tabcolsep}{4pt}
	\begin{center}
		\begin{tabular}{l|c|c}
			\hline 
             Models & ACC & BLEU  \\ \hline 
             SMEA & 76.64 & 24.00   \\
             \hline
             SMEA (w/o memories) & 14.08 & 26.09  \\
			\hline
		\end{tabular}
	\end{center}
	\caption{Ablation test of memory module.}
    \label{tab_ablation}
\end{table}
\begin{table}[tb] 

    \setlength{\tabcolsep}{4pt}
	\begin{center}
		\begin{tabular}{l|c|c}
             \hline 
			 \multicolumn{3}{l}{\text{The staff here is \textcolor{red}{\textbf{very rude}}.}}\\  
			 \multicolumn{3}{l}{\text{It really is n't \textcolor{red}{\textbf{worth coming}} here .}}\\ 
             \multicolumn{3}{l}{\text{\textcolor{red}{\textbf{Very pleased}} with this business.}} \\  
             \multicolumn{3}{l}{\text{Been here once and \textcolor{red}{\textbf{loved}}  going \textcolor{red}{\textbf{here}}.}} \\ 
			\hline
		\end{tabular}
	\end{center}
	\caption{The effectiveness of the memory module with examples. The red words are absent in the input but generated with the help of sentiment memories.}\label{mem_examples}
	\vspace{-0.15in}
\end{table}




\subsection{Error Analysis}

To better interpret our model, we also analyze the failure examples whose sentiments are not transformed. We observe that in most cases, the inputs do not have emotional tendencies. Although we have filtered the sentiment-ambiguous examples in preprocessing, there are still a few ambiguous inputs such as ``What can I say ?'' and ``Been here twice.''. Since our model tries to preserve non-emotional content. These words are easily kept and then the decoder barely depends on sentiment-memories. Thus, it is difficult to handle the sentiment transformation with these examples. 

\section{Conclusion}
In this paper, we propose a model that learns sentiment memories without parallel data and then automatically extract sentiment information to adapt different contexts when decoding. Experimental results show that our method substantially improves content preservation and achieves the state-of-the-art results.

\section*{Acknowledgements}

This work was supported in part by National Natural Science Foundation of China (No. 61673028). We thank all the reviewers for providing the constructive suggestions. Xu Sun is the corresponding author of this paper.

\bibliography{emnlp2018}
\bibliographystyle{acl_natbib_nourl}

\end{document}